# TOWARDS MULTIMODAL CONTENT REPRESENTATION

*Harry Bunt[1] and Laurent Romary[2]*

[1]Computational Linguistics and AI
Tilburg University
P.O. Box 90153
5000 LE Tilburg, the Netherlands
*Harry.Bunt@kub.nl*
let.kub.nl/people/bunt/index.stm

[2]LORIA
Universite de Nancy
B.P. 239
54506 Vandoeuvre-les-Nancy, France
*Laurent.Romary@loria.fr*
www.loria.fr/~romary

## INTRODUCTION

Multimodal interfaces, combining the use of speech, graphics, gestures, and facial expressions in input and output, promise to provide new possibilities to deal with information in more effective and efficient ways, supporting for instance:
- the understanding of possibly imprecise, partial or ambiguous multimodal input;
- the generation of coordinated, cohesive, and coherent multimodal presentations;
- the management of multimodal interaction (e.g., task completion, adapting the interface, error prevention) by representing and exploiting models of the user, the domain, the task, the interactive context, and the media (e.g. text, audio, video).

An intelligent multimodal interface requires a number of functionalities concerning media input processing and output rendering, deeper analysis and synthesis drawing at least upon underlying models of media and modalities (language, gesture, facial expression of user or animated agent),  fusion and coordination of multimodal input and output at a semantic level, interpretation of multimodal input within the current state of the interaction and the context, and reasoning about and planning of multimodal messages. This implies an architecture with many components and interfaces; a reference architecture of an intelligent multimodal dialogue system was established at the workshop `Coordination and Fusion in Multimodal Interaction' in Dagstuhl, Germany, November  2001 (see Bunt, Kipp, Maybury and Wahlster, forthcoming, and http://www.dfki.de/~wahlster/Dagstuhl_Multi_Modality). The communication between many of the components in a multimodal interactive system rely upon an enabling syntax, semantics and pragmatics. A multimodal meaning representation plays central stage in such a system, supporting both interpretation and generation. Such a representation should support any kind of multimodal input and output, and should, in order to be useful in a field which is still developing, be sufficiently open to support a range of theories and approaches to multimodal communication.

The present document is intended to support the discussion on multimodal content representation, its possible objectives and basic constraints, and how the definition of a generic representation framework for multimodal content representation may be approached. It takes into account the results of the Dagstuhl workshop, in particular those of the informal working group on multimodal meaning representation that was active during the workshop (see http://www.dfki.de/~wahlster/Dagstuhl_Multi_Modality, Working Group 4).

## SCOPE

To delineate the task of formulating objectives, constraints and components of multimodal meaning representation, we must first have a shared understanding of what is meant by *meaning* in multimodal interaction. We propose to define the meaning of a multimodal `utterance' as the specification of how the interpretation of the `utterance' by an understanding system should change the system's information state (taken in a broad sense of the term, including domain model, discourse model, user model, task model, and maybe more - see e.g. Bunt, 2000). While formulated with reference to input interpretation only, this definition can also be related to the generation of multimodal outputs, by assuming that an output is generated by the system in order to have an effect on the user through the interpretation of that output by the user. (The generation of appropriate outputs thus depends on the system having an adequate model of what its outputs may mean to the user – which is exactly as it should be.)

A multimodal meaning representation should support the fusion of multimodal inputs and the fission of multimodal outputs at a semantic level, representing the combined and integrated semantic contributions of the different modalities. The interpretation of a multimodal input, such as a spoken utterance combined with a gesture and a certain facial expression, will often have stages of modality-specific processing, resulting in representations of the semantic content of the interactive behavior in each of the separate modalities involved. Other stages of interpretation combine and integrate these representations, and take contextual information into account, such as information from the domain model, the discourse model and the user model. A multimodal meaning representation language should support each of these stages of interpretation, as well as the various stages of multimodal output generation. Since we are considering inputs and outputs from a semantic point of view, the representation of lower-level modality-specific aspects of interactive behavior, like syntactic linguistic information or morphological properties of gestures is not a primary aim, but some such information may percolate as features associated with a meaning representation, especially at intermediate stages of interpretation, where their relevance for semantic interpretation may not have been fully exploited. At the other end of interpretation, where understanding is rooted in information structures like domain models and ontologies, a multimodal meaning representation language should support the connection with frameworks for defining ontologies and specifying domain models, such as DAML + OIL.

While supporting the linking up of meaning representations with ontologies and `low-level' modality-specific information, the design of multimodal meaning representations is to be clearly distinguished from the design of domain model representations, linguistic morphosyntactic representations, representations of facial expressions, etc., which do not fall within this scope. Also, meaning representations should not represent the underlying processes by which they are constructed and manipulated, although it may be important that they are `annotated' with administrative information relating to their processing, such as time stamps.

## OBJECTIVES

The main objective of defining multimodal meaning representations is to provide a fundamental interface format to represent a system's understanding of multimodal user inputs, and to represent meanings that the system will express as multimodal outputs to the user. This interface format should thus be adequate for representing the end result of multimodal input interpretation, and for representing the semantic content that the system will present to the user in multimodal form. It should therefore allow dialogue management, planning and reasoning modules to operate on these representations. In order to be useful for this purpose, this interface format should support the interfaces of these as well as other modules that form part of the system, and thus be adequate not only for representing the end result of semantic interpretation but also intermediate results. Something similar holds for generation. This is a second objective that follows almost immediately from the first.

Another objective in defining a well-defined representational framework for multimodal communicative acts is to allow the specification and comparison of existing application-specific representations (e.g. the M3L representation used in the SmartKom project) and the definition of new ones, while ensuring a level of interoperability between these.

Finally, the specification of a multimodal meaning representation should also be useful for the definition of annotation schemes of multimodal semantic content.

## BASIC CONSTRAINTS

Given the main objective of defining meaning representations, the first and foremost basic requirements of a semantic representation framework are those that we may call `expressive' and `semantic' adequacy:

- *Expressive adequacy:* the framework should be expressive enough to correctly represent the meanings of multimodal communicative acts;
- *Semantic adequacy:* the representation structures should themselves have a formal semantics, i.e., their definition should provide a rigorous basis for reasoning (whether deductive, statistical, in the form of plan operators, or otherwise).

The second objective, of providing interface formats within a multimodal dialogue system architecture, means that `incremental' construction should be supported of intermediate and partial representations, leading up to a final representation or, if the construction of a final representation does not succeed, leading to negative feedback or another appropriate system action. This implies three further basic constraints:

- *Incrementality,* in the sense of supporting various stages of multimodal input interpretation, as well as of multimodal output generation, allowing both early and late fusion and fission;
- *Uniformity:* to make incremental processing feasible, where possible the representation of various types of input and output should be uniform in the sense of using the same kinds of building blocks and the same ways in which complex structures can be composed of these building blocks.
- *Underspecification and Partiality*: to support the representation of partial and intermediate results of semantic interpretation, the framework should allow meaning representations which are underspecified in various ways, and which capture unresolved ambiguities.

Finally, the representational framework should take into account that the design of multimodal human-computer dialogue systems is a developing area in which new research results and new technologies may bring new challenges and new approaches for the representation of multimodal meanings. This means that the representational framework should satisfy the following two constraints:

- *Openness:* the framework should not depend on a single, particular theory of meaning or meaning representation, but should invite contributions from different semantic theories and approaches to meaning representation;
- *Extensibilty.* The framework should be compatible with alternative methods for designing representation schemas (like XML), rather than support only a single specific schema.

## METHODOLOGY

As a first step in the direction of defining a generic multimodal semantic representation form, we have to establish some basic concepts and corresponding terminology.

First, the action-based concept of meaning mentioned above, applicable to multimodal inputs in an interactive situation, means that the meaning of a multimodal `utterance' has two components: one that is often called `propositional' or `referential', and that is concerned with the entities that the utterance refers to and with their properties and relations that may be expressed in propositions, and a `functional' component that expresses a speaker's intention in producing the utterance: what effects does he want to achieve (using `speaker' in a broad, multimodal sense here)? This distinction is familiar from speech act theory, where the two components are called `propositional content' and `illocutionary force', and is also prevalent in other theories of language-based communication (see Bunt, 2000); it is often viewed as drawing a border line between semantics and pragmatics. In the analysis of multimodal interaction it is especially important to pay attention to both these aspects of meaning, since different modalities often contribute to each aspect in different ways; for instance, in spoken interaction the referential and propositional aspects of meaning are often expressed verbally, while gestures and facial expression contribute primarily to the functional aspects. The term `multimodal content' should not be confused with `propositional content', and should not make us forget that multimodal messages have meanings with functional aspects that are equally important as their propositional and referential aspects. In this document we use `multimodal content' as synonymous with `multimodal meaning', including functional aspects, and we use `semantic representation' as synonymous with `representation of meaning'.

A convenient term that has become popular in the literature on human-computer dialogue is `*dialogue act'*. This term is mostly used in an informal, intuitive way, or as a variant of `speech act; it has a formal definition in terms of the effects that a `speaker' intends to achieve through its understanding by the addressee (see Bunt, 2000), which makes it suitably precise for use in the analysis of the meaning of multimodal inputs and outputs. Without further going into definitions here, we will use the term `dialogue act' in the rest of this document. Definitions of other useful concepts can be found in Romary (2002).

As a second methodological step, we propose to distinguish the following three basic types of ingredients that would seem to go into any multimodal meaning representation framework. Each of these ingredients is discussed further in subsequent sections

1. *Basic components*: the basic constructs for building representations of the meaning of multimodal dialogue acts: types of building blocks and ways to connect them.
2. *General mechanisms*: representation techniques like substructure labeling and linking, that make the representations more compact and flexible.
3. *Contextual data categories*: types of administrative (meta-)data that do not, strictly speaking, contribute to the meanings of  semantic representations, but that may nonetheless be relevant for their processing.

## BASIC COMPONENTS

Initially, the following basic components can be identified to represent the general organization of any semantic structure:

1. temporal structures (`*events'*), to represent, for instance:
   - spoken utterances (input or output dialogue acts);
   - gestures (same);
   - noncommunicative action (like searching for information, making a calculation);
   - events, states, processes,.. in the discourse domain, representing meanings of verbs and possibly other linguistic expressions;
2. referential structures (`*participants'*), to represent, for instance:

- the speaker of an input utterance, or the person performing a gesture;
- the addressee of a system output dialogue act;
- individuals and objects participating in a semantic event

3. *restrictions* on temporal and referential structures, to represent, for instance:
   - the type(s) of dialogue, act associated with an utterance;
   - a gesture type, assigned to a gesture token

4. dependency structures, representing *semantic relations* between temporal and/or referential structures, for instance:
   - participant roles (like SPEAKER, ADDRESSEE, AGENT, THEME, SOURCE, GOAL,..)
   - discourse/rhetorical relations
   - temporal relations.

It may be noted that linguistic semantic phenomena that have been studied extensively in relation to the needs of underspecific representation, such as quantification and modification, can also be represented with these basic components. For instance, a quantified statement like `Three men moved the piano' can be represented as a move-event involving a group of three men and a piano, where the collectiveness and the group size of the set of men that form the agent of the event are represented by means of restrictions on the event.

## GENERAL MECHANISMS

In addition to these basic components, certain general mechanisms are important to make meaning representations suitable for representing partial and underspecified meanings, to give the representations a more manageable form, and to relate them to external sources of information. Examples of such mechanisms are:

1. *substructure labeling*: assigning labels to subexpressions and allowing the use of these labels, instead of the substructures that they label, as arguments in other subexpressions;
2. *argument underspecification*: partial or underspecified representations can be constructed using labels in argument positions; restrictions on labels can represent limitations on the ways in which such variables can be instantiated by labels of substructures elsewhere in the representation;
3. *restrictions on label values*: see previous mechanism. Alternatively, *disjunctions*, or *lists* of labels can be used to represent ambiguity or partiality;
4. *structure sharing*, as in typed feature structures, makes it possible to represent that a certain part of the representation plays more than one role, e.g. a participant may be both agent and theme in a semantic event, or may be the speaker of an utterance and the performer of a gesture, as well as the agent in a semantic event expressed by the multimodal dialogue act;
5. *linking to domain models* (types and instances) to anchor meaning representations in the domain of discourse;
6. *linking to lower levels*, such as syntactic structure, prosodic cues, gestural trajectories,.. is useful for tying a purely semantic representation to lower-level information that has given rise to it, and that may not yet have been fully interpreted.

## CONTEXTUAL DATA CATEGORIES

Finally, meaning representations will need to be annotated with general categories of administrative information, both globally and also at the level of subexpressions, to capture certain information which is not found inside the elements of interactive behaviour, but which is potentially relevant for their interpretation and generation, such as:

1. Environment data, for instance:
   - time stamps and spatial information (when and where was this input received, etc.)
2. Processing information, such as:
   - which module has produced this representation; what is its level of confidence, etc.
3. Interactional information:
   - who is the speaker; what addressees are there,..

## Technical Backgound: XML

At this stage, we should say a word about what appear to be the unavoidable technical choices for the definition of a multimodal content representation format that would be used, among other possibilities, to exchange information between processing modules within a man-machine dialogue system. As a matter of fact, f, as defined by the World Wide Web Consortium, appears to be the best candidate so far (and probably for quite a long time) to represent information structures intended to be transmitted across a network. In the following section, we give a very brief overview of XML, which we will then use to illustrate some of the principles mentioned above by means of a concrete example.

XML (eXtended Markup Language) is a simplified (but also in some respects enhanced) version of SGML. It provides a syntax for document markup as well as for the description of the set of tags to be used in classes of documents (a so-called DTD, Document Type Definition). An XML document is made of three parts:

- An XML declaration, which, beyond identifying that the current document is an XML one, allows one to declare the character encoding scheme used in the document (e.g. iso-8859-1, utf-8, etc.);
- A document type declaration, which can point to a DTD. This section can be omitted;
- An XML instance corresponding to the actual data represented by the document.

XML makes an important distinction between a *well-formed* document, which only contains the XML declaration and a syntactically conformant instance, and a *valid* one, where the instance is also checked against the associated DTD.

Among other characteristics, we mention the following important properties of XML:

- XML is both Unicode and ISO 10646 compatible[1]
- XML comes along with a specific mechanism, called *namespaces*, allowing one to combine, within the same document, markup taken from multiple sources. This very powerful mechanism, which is in particular the basis for XSLT and XML schemas, allows more modularity in the definition of an XML structure and also to reuse components defined in another context;
- XML provides a general attribute 'xml:lang' to indicate the language used in a given element (see above).

The W3C also provides three very important recommendations for traversing XML documents, namely:

- XPath, which describes a syntax and associated mechanisms to move within a document instance;
- XPointer, which allows one to indicate a location within a document and is based upon the XPath recommendation;
- XLink, which allows one to combine and qualify a set of pointers to describe a link between them.

These three recommendations are important for instance when one wants to relate some information produced by a given processing level and the information that has been used as input for those processes.

Still, it should be noticed that the existence of such a widely recognized *metalanguage* as XML does not solve our problems for representing multimodal content. First, XML by itself does not come with a formal semantics for its tags, and thus does not satisfy the requirement of semantic adequancy. Second, the requirements of flexibility and extensibility forbid us to try to standardize once and for all a precise XML format, but rather think of providing concepts and tools for anyone to be able to design his or her own format, while preserving interoperability conditions with someone else's choices. This is the spirit in which work has already been done within TC37/SC4 for the definition of TMF (Terminological Markup Framework; ISO 16642, under DIS ballot), first proposed for lexical information by Ide, Kilgariff and Romary (2000) and which has recently been taken over to deal with morphosyntactic and syntactic annotation (see (Ide & Romary, 2001a and Ide & Romary, 2001b, respectively).
The basic assumption that we make is that there exists an entire class of document formats that can be modelled by combining a *metamodel*, that is an abstract structure shared by all documents of a given type (e.g. syntactic annotation document), with a choice of the data categories that may be associated with the various levels of the metamodel. Such a description can be seen as a specification of the document format, which can be instantiated by providing XML representations for the metamodel and the data categories. In such a view, if a community of researchers and implementers agrees on the definition of a reduced set of metamodels for language resources, the actual choice of data categories is left to the responsibility of a specific application. In this framework, the interoperability between formats is ensured by providing a data category registry which gathers, together with precise reference and definition, the various data categories needed for a particular field.

In the case of multimodal content representation we thus advocate that, beyond agreement on the basic components and mechanisms for instance as described in this paper, which could go into the definition of an actual metamodel for content representation, one should not try to standardize a particular XML format more precisely (though we need to make specific choices to illustrate our approach with concrete examples, see below).

**A simple example**
In the following, we illustrate the possible combination of basic components, general mechanisms, and contextual data categories into a multimodal meaning representation. This representation exemplifies the general methodology that we suggested here, by taking up a sample semantic representation derived from an initial example expressed in the ULF+ format (ULF+ is a slightly updated version of a semantic representation language that was developed successively in the PLUS dialogue project, see Geurts and Rentier, 1993, and in the multimodal DENK project; see Bunt et al., 1998; Kievit, 1998).

---

[1] The W3C has put pressure on both ISO and the Unicode consortium to make sure that they would not diverge in their parallel work on the definition of a universal character encoding scheme.

In the XML excerpt below (corresponding to the sentence "I want to go from Paris to Stuttgart" uttered by a speaker named Peter), we have extended the original ULF+ representation to introduce the notion of dialogue act, whose participants are the speaker and the system. This example is intended to show how we can differentiate between three types of information in such a representation:

- The instantiation of the semantic content representation metamodel as an XML outline (shown in <u>underlined characters</u>), which organizes the general information layout of the data to be represented;
- The actual information units describing the various levels in the XML outline (shown in gray characters);
- The generic mechanisms used to combine events, participants, restrictions and relations (indicated in **bold characters**).

The specific choices made in this example to represent the metamodel or the data categories as XML objects are only one possibility among many, and this does not affect the formal semantics of the underlying information structure. More precisely, the following explanation may help to clarify the example:

- The <semRep> element corresponds to the semantic representation of one elementary utterance or dialogue act. It is identified uniquely by an id attribute;
- The <event> element is used in this example to represent both the dialogue act proper ("e1") and the event expressed by the corresponding linguistic content ("e2");
- The <participant> element is used to represent the various entities involved in the events. Events and participants being related to one another by means of <relation> elements (with source and target attributes pointing to the corresponding arguments of the relation).

The various levels are then further described by a number of data categories, chosen here to illustrate the wide variety of possible cases. Notice the use of an <alt> structure to illustrate the case where an ambiguity would remain at a given step of analysis, each possibility being associated with a certainty evaluation ('cert' attribute). In accordance with the methodology developed in TMF, the name of the corresponding XML elements and attributes should not be the object of standardization, data categories being defined by abstract properties.

```
<semRep id="rep1">
    <event id="e0">
      <evtCat>utterance</evtCat>
      <speaker target="Peter"/>
      <adressee target="System"/>
      <alt>
          <dialAct cert="0.8">Order</dialAct>
          <dialAct cert="0.3">Inform</dialAct>
      </alt>
    </event>

    <participant id="Peter">
      <!-- A description of the speaker that can be referendum
      elsewhere in the document -->
    </participant>

    <event id="e1">
      <tense>present</tense>
      <voice>active</voice>
      <wh>none</wh>
      <evtType>wanttogo</evtType>
      …
    </event>

    <participant id="x">
      <lex>I</lex>
      <synCat>Pronoun</synCat>
      <num>sing</num>
      <pers>first</num>
      …
    </participant>
```

```
    <participant id="y">
      <lex>Nancy</lex>
      <synCat>ProperNoun</synCat>
      <pers>third</num>
      …
    </participant>

    <participant id="z">
      <lex>Stuttgart</lex>
      <synCat>ProperNoun</synCat>
      <pers>third</num>
      …
    </participant>

    <relation source="x" target="e1">
      <role>agent</role>
    </relation>

    <relation source="y" target="e1">
      <role>source</role>
    </relation>

    <relation source="y" target="e1">
      <role>goal</role>
    </relation>
</semRep>
```

## ACTION PLAN

The variety of existing theoretical approaches, as well as the wide number of factors to be considered makes it very difficult to devise from scratch a truly generic framework for multimodal content representation. As a consequence it is necessary to involve, beyond the possibilities offered by the definition of a working group on this topic in TC37/SC4, as large a community of experts as possible in the development of such a framework. This is why we suggest that the work shall be initially conducted within a dedicated working group of SIGSEM (Special Interest Group on Computational Semantics of the Association of Computational Linguistics), which would be, right from the beginning, a liaison with TC37/SC4. This group would prepare a working draft, which would then be submitted to ISO.

Doing so, it would also be easier to ensure a proper interaction with other interested communities, in particular the people working on multimedia representation (SIGMedia, in complement to the existing liaison between MPEG and TC37/SC4) and on discourse and dialogue (SIGDial).

The agenda would thus be the following:

- Refining the workplan on the basis of the present paper  at the TC37/SC4 Preliminary Meeting in Jeju (Korea) in February 2002.
- Presenting a position paper at the LREC workshop on "International Standards of Terminology and Language Resources Management" in May 2002.
- First working group meeting in conjunction to IWCS-5 (5th International Workshop on Computational Semantics) in Tilburg, the Netherlands, in January 2003.